\documentclass{applemlr}

\usepackage{amsmath}
\usepackage{enumerate}
\usepackage{algorithm}
\usepackage{algpseudocode}
\usepackage{amsfonts}
\usepackage{amsthm}
\usepackage{cleveref}
\usepackage{diagbox}
\usepackage{colortbl}
\usepackage{amssymb}
\usepackage{xspace}
\usepackage{wrapfig}
\usepackage{adjustbox}
\usepackage{tabularx}
\usepackage{booktabs}
\usepackage{mathtools}
\usepackage{tikz}
\usepackage{enumitem}
\usepackage{silence}
\usepackage{dsfont}
\usepackage{makecell}
\usepackage{xfakebold}
\usepackage{multirow}

\usepackage{amsmath,amsfonts,bm}

\def\eqref#1{equation~\ref{#1}}

\def\1{\bm{1}}

\DeclareMathAlphabet{\mathsfit}{\encodingdefault}{\sfdefault}{m}{sl}
\SetMathAlphabet{\mathsfit}{bold}{\encodingdefault}{\sfdefault}{bx}{n}

\definecolor{textgray}{HTML}{6E6E73}
\usetikzlibrary{positioning, calc}
\usetikzlibrary{decorations.pathmorphing}

\makeatletter
\patchcmd{\wrong@fontshape}{\@gobbletwo}{}{}{}
\makeatother
\numberwithin{equation}{section}
\makeatletter
\AtBeginDocument{
  \urlstyle{sf}
  
}
\makeatother

\definecolor{light}{RGB}{125, 125, 125}
\crefname{tcb@cnt@pbox}{code}{code}
\Crefname{tcb@cnt@pbox}{Code}{Code}
\crefname{assumption}{assumption}{assumption}
\Crefname{assumption}{Assumption}{Assumptions}

\newtcolorbox[auto counter]{pbox}[2][]{
  colback=white,
  title=Code~\thetcbcounter: #2,
  #1,fonttitle=\sffamily,
  fontupper=\sffamily,
  arc=2pt,
  colframe=bgcolor,
  coltitle=fgcolor,
  colbacktitle=bgcolor,
  toptitle=0.25cm,
  bottomtitle=0.125cm
}

\makeatletter
\newcommand\applefootnote[1]{%
  \begingroup
  \renewcommand\thefootnote{}%
  \renewcommand\@makefntext[1]{\noindent##1}%
  \footnote{#1}%
  \addtocounter{footnote}{-1}%
  \endgroup
}
\makeatother

\definecolor{cverbbg}{gray}{0.90}

\title{Hybrid Modeling of Photoplethysmography for Non-invasive Monitoring of Cardiovascular Parameters}

\author[1,*]{Emanuele Palumbo}
\author[2]{Sorawit Saengkyongam}
\author[2]{Maria R. Cervera}
\author[2]{Jens Behrmann}
\author[2]{Andrew C. Miller}
\author[2]{Guillermo Sapiro}
\author[2]{Christina Heinze-Deml}
\author[2]{Antoine Wehenkel}

\affiliation{$^1$ETH Zurich}\affiliation{$^2$Apple}

\abstract{
Continuous cardiovascular monitoring can play a key role in precision health.  However, some fundamental cardiac biomarkers of interest, including stroke volume and cardiac output, require invasive measurements, e.g., arterial pressure waveforms (APW). As a non-invasive alternative, photoplethysmography  (PPG) measurements are routinely collected in hospital settings.
Unfortunately, the prediction of key cardiac biomarkers from PPG instead of APW remains an open challenge, further complicated by the scarcity of annotated PPG measurements. 
As a solution, we propose a hybrid approach that uses hemodynamic simulations and unlabeled clinical data to estimate cardiovascular biomarkers directly from PPG signals. Our hybrid model combines a conditional variational autoencoder trained on paired PPG-APW data with a conditional density estimator of cardiac biomarkers trained on labeled simulated APW segments. As a key result, our experiments demonstrate that the proposed approach can detect fluctuations of cardiac output and stroke volume and outperform a supervised baseline in monitoring temporal changes in these biomarkers.}

\metadata[Correspondence]{\sffamily Emanuele Palumbo: \url{emanuele.palumbo@inf.ethz.ch}; Antoine Wehenkel: \url{awehenkel@apple.com}}
\date{\sffamily\today}

\makeatletter
\newcommand{\blfootnoterb}[1]{%
  \begingroup
    \renewcommand\thefootnote{}\footnote{#1}%
    \addtocounter{footnote}{-1}
  \endgroup
}
\makeatother

\begin{document}

\maketitle

\begin{figure}[ht]
    \centering
    \includegraphics[width=1.0\linewidth]{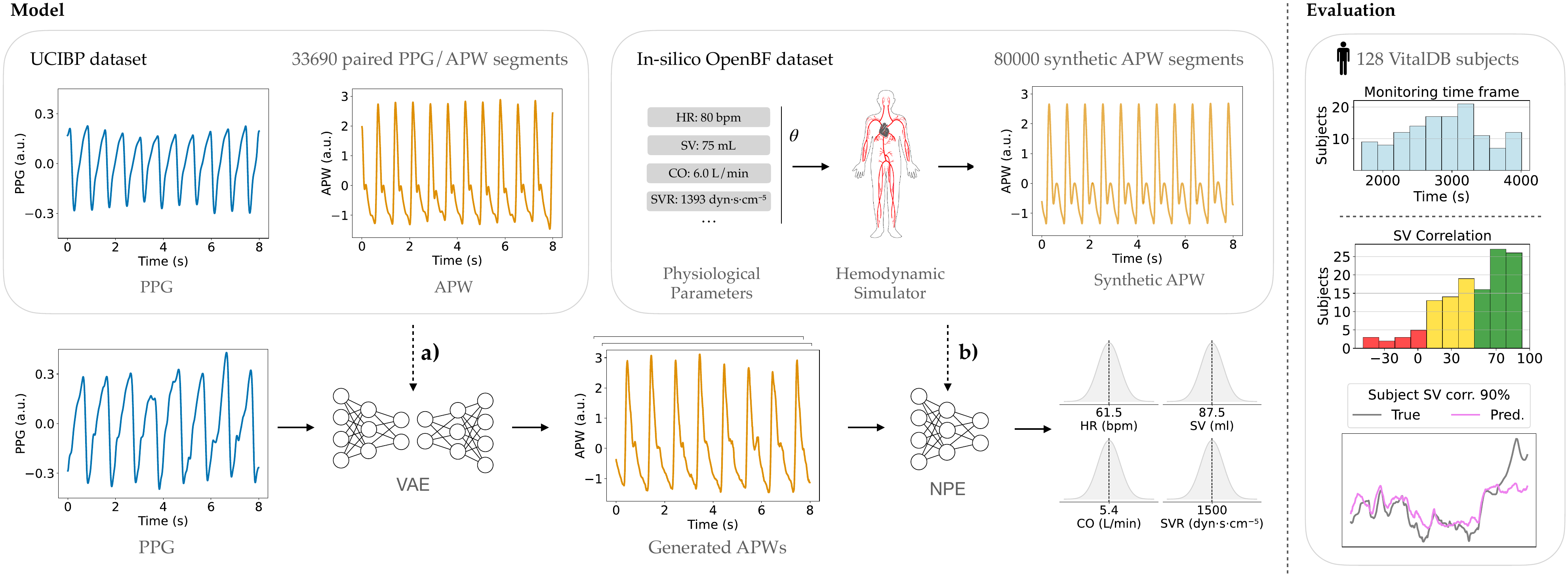}
    \caption{\small Proposed hybrid framework for non-invasive monitoring of cardiovascular parameters from PPG signals. The framework combines simulated and real-world data across two key components: \textbf{a)} a conditional VAE, trained on real-world paired APW and PPG segments, producing plausible APW signals from PPG inputs; \textbf{b)} a density estimator trained on a in-silico dataset of hemodynamic simulations inferring  cardiovascular biomarkers from generated APW signals. The framework is validated in-vivo on unseen VitalDB data, using per-subject Spearman correlation between ground truth and predicted values of cardiovascular biomarkers.}
    \label{fig:approach-overview}
\end{figure} 
\vspace{-1ex}
\blfootnoterb{$^*$ Work done during an internship at Apple.}

\section{Introduction}
\label{sec:intro}

Continuous monitoring of cardiovascular parameters such as cardiac output~(CO) and stroke volume~(SV) has proven essential to evaluate function, guide treatment, and detect hemodynamic instability  in critical care \citep{PINSKY2007Hemodynamic, Nguyen2017noninvasive}. However, current clinical assessment relies on invasive techniques that entail risk and require medical supervision \citep{Joosten2017Accuracy, Arya2022COMonitoring, evans2009Complications}.
Photoplethysmography (PPG), which captures cardiovascular information non-invasively and is increasingly accessible through consumer wearables \citep{Charlton2024WearablePPG, Weng2024PredictingCV}, offers a promising alternative that could enable safer monitoring in critical care and potentially extend capabilities beyond hospital settings \citep{Orini2023Premature, Williams2023Wearable}. Yet, monitoring cardiac biomarkers such as SV and CO from PPGs remains challenging due to the scarcity of labeled data and the indirect, not fully understood relationship between PPG signals and these parameters.

In the context of cardiovascular monitoring, physiological simulations \citep{charlton2019modeling, Melis2017Bayesian} emerge as a potential strategy to mitigate the scarcity of in-vivo (i.e., real-patient) labeled data with cardiovascular biomarker annotations, using in-silico (i.e., simulated) data. In particular, recent work demonstrated the value of hemodynamic simulators to obtain arterial pressure waveform (APW) signals for inferring cardiovascular biomarkers in in-vivo data \citep{Manduchi2024Leveraging}. While closely related to PPG signals, APWs are obtained invasively and thus provide a more direct view of central hemodynamics. By contrast, PPG reflects peripheral vascular changes through a complex relationship influenced by exogenous factors, including temperature, skin properties, and motion artifacts \citep{meier2024wildppg}. Windkessel models attempt to describe the APW–PPG relationship, but they are too constrained to capture the full complexity of real-world data. Consequently, \citet{Manduchi2024Leveraging} reported mixed results when applying the same strategy to PPG, despite its effectiveness on APWs.

To address these limitations, we propose a novel hybrid learning strategy that combines hemodynamic simulations with in-vivo data to estimate cardiac biomarkers directly from PPG signals. As in \citet{Manduchi2024Leveraging}, we first use a simulator of APWs to train a neural posterior estimator ~\citep[NPE,][]{lueckmann2017flexiblestatisticalinferencemechanistic}  to model the relationship between APWs and cardiovascular parameters. To learn the transfer function between APW and PPG modalities, we then train a generative model on a real-world dataset of paired PPG-APW signals. Taken together, our contributions are: \textbf{(i)} a novel hybrid approach leveraging both in-silico simulations and in-vivo biosignals for PPG-based cardiovascular assessment;  \textbf{(ii)} validation on unseen in-vivo data for the monitoring of stroke volume and cardiac output changes; and  \textbf{(iii)} a demonstration that physics-based simulation enables robust cardiovascular parameter estimation from PPG signals.

\section{Method}
\label{sec:methodology}
Our goal is to predict some cardiovascular parameters of interest given a finger PPG measurement, for which we propose a novel approach illustrated in \Cref{fig:approach-overview}. We formalize this task as learning a conditional density estimator of $p(\theta \mid \mathbf{y})$, where $\mathbf{y} \in \mathbb{R}^T$ denotes a T-timestep PPG measurement and $\theta \in \mathbb{R}^{d}$ the parameters of interest. To solve this task, we are given \textbf{1.} a large dataset of labeled simulated APWs $\mathcal{D}_s := \{\theta_i, \mathbf{x}_i\}_{i=1}^{N_s}$, where $\theta_i \sim \pi(\theta)$ are sampled from a prior distribution $\pi$, and $\mathbf{x}_i \sim \tilde{p}(\mathbf{x} \mid \theta = \theta_i)$\footnote{The notation ``$\tilde{p}$ '' emphasizes that the simulator approximates the real world.} are generated by running the simulator on $\theta_i$; and \textbf{2.} a dataset $\mathcal{D}_r = \{\mathbf{x}_i, \mathbf{y}_i\}_{i=1}^N$ of simultaneous real-world APW and PPG measurements. In addition, for evaluation, we use a labeled dataset $\mathcal{D}_{e} := \{\theta_i, \mathbf{y}_i\}_{i=1}^{N_e}$ containing PPGs paired with ground truth parameter values. 

\paragraph{Estimating $p(\theta \mid \mathbf{y})$.}
We approximate $p(\theta \mid \mathbf{y})$ by ignoring direct dependencies between parameters $\theta$ and PPG measurements $\mathbf{y}$, leading to the approximation $$p(\theta \mid \mathbf{y}) \approx \mathbb{E}_{p(\mathbf{x} \mid \mathbf{y})}[p(\theta \mid \mathbf{x})].$$
We then learn two estimators, $p_{\phi_s}(\theta \mid \mathbf{x})$ and $p_{\phi_r}(\mathbf{x} \mid \mathbf{y})$ on the labeled simulation dataset $\mathcal{D}_s$ and paired APW-PPG dataset $\mathcal{D}_r$, respectively. 

\paragraph{Deep generative modeling of $p(\mathbf{x} \mid \mathbf{y})$ from $\mathcal{D}_r$.}
We train a conditional variational auto-encoder~\citep[VAE,][]{kingma2014autoencoding} on $\mathcal{D}_r$ to estimate $ p(\mathbf{x} \mid \mathbf{y})$, the mapping from PPG to APW segments. In particular, we assume a latent generative model $p_{\phi_r}(\mathbf{x} \mid \mathbf{y}) := \mathbb{E}_{p(\mathbf{z})}[p_{\phi_r}(\mathbf{x} \mid \mathbf{z}, \mathbf{y})]$, where the latent $\mathbf{z}\sim p(\mathbf{z}) := \mathcal{N}(0, I)$ aims to capture unobserved physiological variations impacting the relationship between APWs and PPGs. In our experiments, we use 1D-CNN architectures and Gaussian densities to parameterize the encoder $q_{\phi_r}(\mathbf{z} \mid \mathbf{x}, \mathbf{y})$ and decoder $p_{\phi_r}(\mathbf{x} \mid \mathbf{z}, \mathbf{y})$ networks. We train these two networks by optimizing the Evidence Lower Bound objective on the the UCI Cuff-Less Blood Pressure Estimation (UCIBP) dataset \citep{UCIBPpaper}, containing $333\text{,}690$ paired APW and PPG 8-second segments from ICU patients. We provide details on the data format, architecture and training settings in \Cref{app:exp_details_VAE}. 

\paragraph{Neural posterior estimation of  $p(\theta \mid \mathbf{x})$.}
Following \citet{Manduchi2024Leveraging}, 
we train a neural posterior estimator (NPE) $p_{\phi_s}(\theta \mid \mathbf{x})$ by optimizing directly $\sum_{i=1}^{N_s} \log p_{\phi_s}(\theta_i \mid \mathbf{x}_i)$ via stochastic gradient ascent. To make $p_{\phi_s}(\theta_i \mid \mathbf{x}_i)$ differentiable we parameterize it with a conditional normalizing flow~\citep[NF,][]{rezende2016variationalinferencenormalizingflows} composed of $3$ steps of autoregressive and affine transformations and a convolutional neural network that encodes the APW. For our experiments, we reproduce the dataset from \citet{Manduchi2024Leveraging}, which contains $\approx 32\text{,}000$ synthetic APW segments paired with parameter values. \Cref{app:exp_details_NPE} further documents how we train the NPE model.

\paragraph{Forming predictions with $p_{\phi}(\theta \mid \mathbf{y})$.}
For a given PPG segment $\mathbf{y}$, we first sample $M$ plausible corresponding APW segments $\mathbf{x}_m \sim p_{\phi_r}(\mathbf{x} \mid \mathbf{y}, \mathbf{z}_m)$, where $\mathbf{z}_m \sim \pi$. Then, from each corresponding estimated posterior distribution $p_{\phi_s}(\theta \mid \mathbf{x}_m)$ 
we draw $K$ samples. To obtain a point estimate we average the obtained $M \times K$ samples, and their standard deviation is used as a measure of uncertainty.

\section{Results}
\label{sec:experiments}

We evaluate our approach on an entirely separate dataset, namely the VitalDB dataset \citep{Lee2022VitalDBAH}, comprising APW and PPG signals from $128$ patients undergoing non-cardiac surgery, labeled with cardiovascular biomarkers. Our evaluation focuses on four key cardiovascular parameters: Heart Rate (HR), Stroke Volume (SV), Cardiac Output (CO), and Systemic Vascular Resistance (SVR). As a quantitative metric we report the per-patient Spearman correlation between ground truth and predicted values for a given cardiovascular parameter over the time frame that a patient is monitored. 

\begin{figure}[htp]
    \centering
\includegraphics[width=1.0\linewidth]{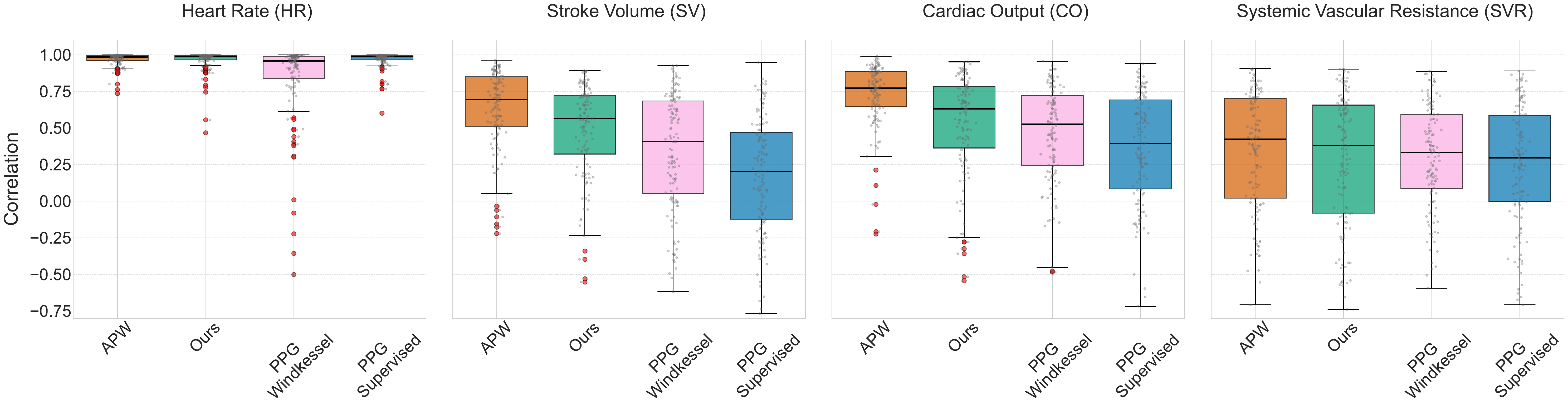}
\caption{ \small
Box plots illustrating the per-subject Spearman correlation coefficients for the estimation of four key cardiovascular parameters. Our proposed hybrid approach (green) is compared against three relevant baselines: (i) \textit{APW} (orange), which uses the ground-truth invasive APWs for inference with estimator trained in-silico; (ii) \textit{PPG Windkessel} (pink), an estimator trained on labeled in-silico PPGs obtained with a Windkessel model approximation~\citep{Manduchi2024Leveraging}; and (iii) \textit{PPG Supervised} (blue), an estimator trained directly on a subset of labeled VitalDB PPG data. Average results across independent runs with standard deviations are reported in \Cref{tab:ext-num-results}. 
} 
\label{fig:cv-params-boxplots}
\end{figure}

\paragraph{Usefulness of hybrid modeling.}
\Cref{fig:cv-params-boxplots} compares our hybrid approach with three relevant baselines, described in the caption. First, all methods perform well at predicting HR. This confirms that our pipeline is capable, in principle, of predicting cardiovascular parameters whose effect is faithfully captured by the simulator. More interestingly, our method achieves a median per-subject correlation above $0.5$ for both SV and CO, significantly higher than what \citet{Manduchi2024Leveraging} obtained using a prescribed Windkessel transfer function between APWs and PPGs. Directly using groundtruth APWs as input to the predictive model, as in the APW baseline, expectedly outperforms our method, which instead relies on APW estimates. However, our method outperforms a PPG-supervised baseline in this task despite not having seen any in-vivo data with cardiovascular parameter labels, in contrast to this baseline. This comparison underscores the value of hemodynamic simulations for estimation of cardiovascular biomarkers from PPG inputs, and the effectiveness of combining prescribed simulations with learned components. Unlike purely model-based approaches, which often suffer from oversimplified assumptions (such as the Windkessel model), our hybrid modeling strategy has the potential to mitigate the absence of large labeled datasets.

\paragraph{Usefulness of generative modeling.}
We demonstrate that learning the PPG-to-APW transfer function with a conditional generative model offers benefits over deterministic models for the downstream tasks we consider in the experiments. Estimating APW from PPG is inherently difficult because PPG signals reflect peripheral vascular changes and are significantly influenced by noise, motion, and other confounding factors, making their relationship to invasive APWs complex and indirect. Hence deterministic models must compromise on this ambiguity, leading to a suboptimal information bottleneck for the parameters of interest. \Cref{tab:vitaldb_spearman_correlation} in \Cref{app:gen_model} compares our method with a deterministic model and shows that deterministic models consistently achieve suboptimal performance. Furthermore, \Cref{fig:uncertainty-PPGs-combined} shows that prediction uncertainty tracks signal quality, highlighting the value of generative models for uncertainty modeling.

\begin{figure}[htp]
    \centering
    \includegraphics[width=.8\linewidth]{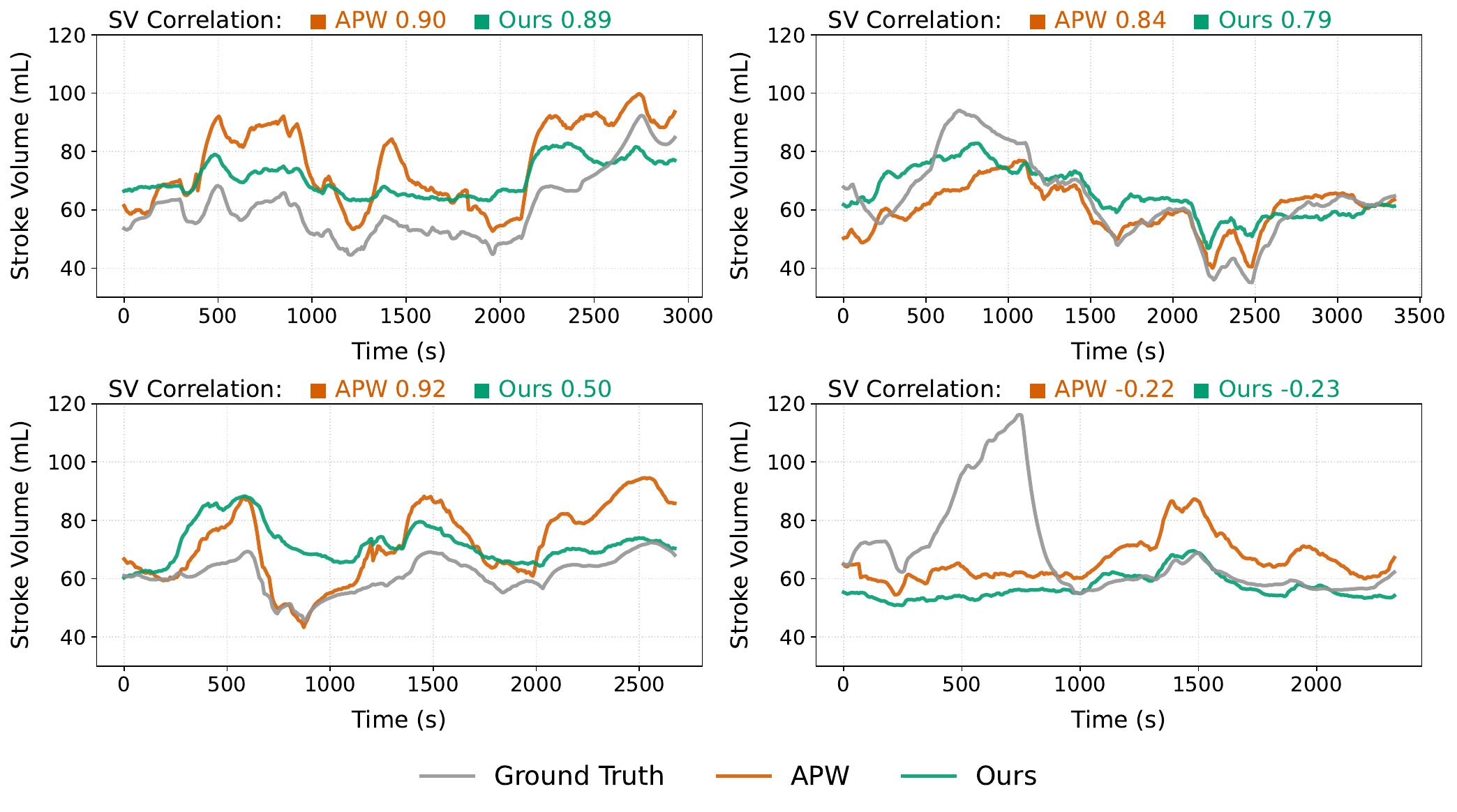}
    \caption{\small SV ground truth (gray), predicted values with our approach (green), and redictions obtained with invasive inference from ground truth APW (orange), on different subjects from the VitalDB dataset.}
    \label{fig:per-subject-sv-monitoring}
\end{figure}

\paragraph{On the difficulties of absolute value prediction.}
In \Cref{fig:per-subject-sv-monitoring}, we examine the SV predictions for a subset of patients. While our approach reliably captures temporal trends in SV (e.g., it captures SV and CO peaks accurately), predicting the absolute values of such complex hemodynamic parameters remains a persistent challenge \citep{Manduchi2024Leveraging}. Our results in \Cref{tab:MAE-results} confirm this observation, showing that our method, as the other compared methods in this work, achieves limited effectiveness for absolute value prediction. This challenge stems from a combination of factors: (i) our current estimator performs a per-segment prediction and lacks a notion subject, thus overlooking unobservable, subject-specific physiological characteristics that influence baseline SV; and (ii) despite its sophistication, the underlying hemodynamic simulator may have inherent model misspecification when applied to real-world data. 
As future work, predictive model calibration on a small set of labeled data shall improve absolute value prediction. Furthermore, enabling personalized predictions, e.g., by introducing a notion of subject in the simulations and via the PPG-to-APW model, is another promising avenue.

\section{Conclusion}
\label{sec:conclusion}

In this work we use a hybrid modeling approach to infer cardiovascular parameters from in-vivo PPG signals. Compared to purely data-driven approaches that struggle due to limited labeled data, our method achieves promising results by incorporating simulations and sidestepping the need for invasive and costly annotations. While other existing hybrid approaches for cardiovascular modeling either embed physical properties as structural constraints within neural networks \citep{Sel2023, zhang2025physiologicalmodelbasedneuralnetworkframework} or augment traditional physiological models with data-driven components \citep{nazaret2023}, our method incorporates physical knowledge in the model through SBI. While SBI relies on accurate forward models, in practice these may not be available to fully characterize each aspect of a complex process. We address this challenge and integrate a data-driven generative model which approximates the PPG-to-APW mapping, for which an accurate physics-based characterization is currently lacking. Our results contribute to characterizing the informativeness of PPG signals for predicting cardiac biomarkers, and could extend beyond the ones considered in our experiments. While our results are promising in monitoring temporal trends, absolute value prediction of complex biomarkers remains challenging, and is a key direction for future work. Future work may also explore alternative generative approaches for the PPG-to-APW mapping, or investigate different architectural choices. Finally, a similar learning strategy than the one used here for finger PPG could extend to other modalities, including wearable PPG, and open the door to passive and long-term cardiac biomarker monitoring.

\bibliographystyle{plainnat}
\bibliography{jmlr_sample}

\clearpage

\appendix

\section{Experimental details}\label{apd:exp_details}

\subsection{Datasets details}
\label{apd:datasets}
In this section we provide additional details for the in-silico and in-vivo datasets used in this work.
It is worth noting that all PPG datasets considered in this work consist of finger PPG signals.

\paragraph{In-silico datasets} We utilize an in-silico dataset of simulated Arterial Pressure Waveforms (APWs) generated using the OpenBF simulator, an open-source 1D blood flow solver \citep{Benemerito_2024}, recently introduced by \citet{Manduchi2024Leveraging}. This dataset consists of 8-second 125Hz-frequency APW segments, synthesized from a comprehensive set of physiological parameters, that characterize the cardiovascular state for 80,000 virtual subjects. These parameters characterize heart function (e.g. Heart Rate, Stroke Volume, Peak Flow Time), arterial properties (e.g. arteries diameter/length) and vascular beds. The simulation process incorporates a stochastic noise model, combining Gaussian and red noise, to mimic the noise  present in real measurements. Finally, the generated signals undergo a bandpass filtering. 
For the comparisons in \Cref{sec:experiments} we use a PPG in-silico dataset, also introduced in \citet{Manduchi2024Leveraging}. Similar to APW generation, cardiovascular states for 80,000 patients are characterized by a comprehensive set of physiological parameters, and the PPG model is modeled by a simplistic Windkessel model \citep{Manduchi2024Leveraging, charlton2019modeling}. While 80,000  APW and PPG signals were generated for both in-silico datasets, samples with anomalous systolic and diastolic bloop pressure were filtered out, leaving $\approx 32,000$ waveforms after this filtering. 

\paragraph{In-vivo datasets}
We use two different in-vivo datasets in this work. The UCI Cuff-Less Blood Pressure Estimation (UCIBP) dataset \citep{UCIBPpaper} provides paired APW and PPG 8-second segments from ICU patients in the MIMIC-II dataset \citep{MIMIC-IIpaper}, without subject information. The signals are sampled at 125Hz frequency. Note that the biosignals in this dataset are not annotated with any cardiovascular labels. For our evaluations in \Cref{sec:experiments} we use the VitalDB dataset, which includes intraoperative APW and PPG signals from patients undergoing non-cardiac surgery in the Seoul National University Hospital with associated cardiovascular parameters. As cardiovascular biomarker labels we consider Heart Rate (HR), Stroke Volume (SV), Cardiac Output (CO), and Systemic Vascular Resistance (SVR). As in \cite{Manduchi2024Leveraging}, we process the VitalDB data to remove faulty signals (e.g. containg NaNs or implausible minimum or maximum values). We split the resulting APW and PPG waveforms in 8-second segments to match the lenght of simulated/UCIBP data. 

\subsection{Model and Implementation Details}
\subsubsection{Conditional VAE model}\label{app:exp_details_VAE} We use 1D-CNNs to parameterize the encoder and decoder of our conditional VAE model. We split the UCI Cuff-Less Blood Pressure Estimation (UCIBP) dataset as follows. We use $333,690$  paired APW-PPG segments from this dataset, randomly split in training ($80 \%$) and validation ($20\%$) sets. We set the latent space size of the VAE to 128 dimensions. The VAE model is trained using the Adam Optimizer by maximizing the ELBO objective referenced in \Cref{sec:methodology} for $300$ epochs, with a learning rate of $0.0005$, batch size of $256$, and early stopping on the validation dataset.  

\paragraph{Encoder architecture}
The encoder processes PPG and APW 
8-second segments at 
125Hz frequency ($\in \mathbb{R}^{1000}$), which are concatenated to obtain a two-channel input. This combined input passes through a sequence of four 1D convolutional layers, each with a kernel size of 5, stride of 2, and padding of 2. The output channels are set to 32, 64, 128, and 256 for the respective layers. Each convolutional layer is followed by Batch Normalization. From the flattened output of this block of convolutional layers two distinct heads compute the mean and log-variance of the encoding distribution.

\paragraph{Decoder architecture}
The decoder takes as input a PPG segment $\mathbf{y}$
and the latent embedding $\mathbf{z}$.
It incorporates a feature extractor network, which processes the 8-second, 125Hz frequency PPG segment ($\in \mathbb{R}^{1000}$) through a sequence of four 1D convolutional layers. Each of these layers uses a kernel size of 5, stride of 2, and padding of 2, and is followed by a LeakyReLU activation. The output channels of these four layers are set to 16, 32, 64, and 128 respectively. The flattened output from this block of  convolutional layers is then projected via a dedicated head to the dimensionality of the latent embedding $\mathbf{z}$, and concatenated with the latent embedding $\mathbf{z}$. This combined vector is transformed by a linear layer, which prepares it to be input to a block of four 1D transposed convolutional layers, beginning with 256 input channels. Each of these layers has a kernel size of 5, stride 2, padding 2, and output padding 1, and is followed by Batch Normalization and LeakyReLU. The output channels for the first three layers are set to 128, 64, 32, while the last layer has a single output channel. The resulting waveform is then cropped to 1000 time steps.

\subsubsection{NPE model} \label{app:exp_details_NPE} To invert the forward process of the OpenBF APW simulator, mapping $\theta \in \mathbb{R}^d$ to a given APW segment $\mathbf{x} \in \mathbb{R}^T$, we follow recent work \citep{Manduchi2024Leveraging} and train a Neural Posterior Estimator (NPE) to approximate the posterior distribution $p(\theta \mid \mathbf{x})$ via normalizing flows \citep{rezende2016variationalinferencenormalizingflows}. Following \citet{Manduchi2024Leveraging} we use autoregressive normalizing flows, thereby minimizing the loss

\begin{align*}
\ell_{\text{NPE}}(\phi_s) = & \frac{1}{N_s} \sum_{i=1}^{N_s} \log p_z \left(f_{\phi_s}(\theta_i;\mathbf{x}_i)\right) 
+ \log \left|J_{f_{\phi_s}}(\theta_i;\mathbf{x}_i)\right|,
\end{align*}
where $p_z$ denotes the isotropic Gaussian distribution and $J_{f_{\phi_s}}$ is the Jacobian's determinant of the function $f_{\phi_s}$. Note that the function $f_{\phi_s}(\theta_i;\mathbf{x}_i)$ decomposes as $f_{\phi_s}(\theta_i;\mathbf{x}_i) = h_\tau(\theta_i;g_{\eta}(\mathbf{x}_i))$ where $h_\tau$ is invertible with respect to the first argument and parameterized via a three-step autoregressive normalizing flow, and $g_{\eta}$ is a 1D-CNN encoder processing the APW segment into a 140-dimensional representation, used to condition the normalizing flow.  The 1D-CNN encoder consists of a block of five 1D convolutional layers with kernel size 3, stride 2 and no padding, each followed by ReLU activation, interleaved by a MaxPool layer with kernel size 2 between the third and fourth layer. The output channels of these layers are set to 40, 40, 40, 20, 10. The output from this block is flattened, and fed to a two-layer MLP head mapping to a 140-dimensional representation. The NPE model is trained for 50 epochs with a learning rate of 0.0003 and batch size $256$, with early stopping on a validation set ($20\%$). At inference time, we sample $K=20$ values for the $\theta$ parameters from the normalizing flow model for each of $M=10$ APW generations obtained with the conditional VAE model (see also \Cref{sec:methodology}).

\subsubsection{Baselines}

In this section we describe implementation and training details for the baselines included in \Cref{sec:experiments}. We report three key baseline comparisons. 
\paragraph{APW:} We compare our approach with inference performed directly on ground truth VitalDB APWs, using the same NPE model trained on in-silico APWs that we use as part of our proposed approach. Note that in comparing with this baseline one should keep in mind that it requires invasive measurements. We train the model  for 50 epochs with a learning rate of 0.0003 and early stopping on the validation set.
\paragraph{PPG Supervised:} We also include as a baseline a supervised density estimator trained on a subset of VitalDB subjects. For consistency, we use the same architecture as the NPE model trained on in-silico APW that is part of our approach. We train the model on the labeled PPG data for 86 (two thirds) of the patients for training, and split the remaining ones equally between validation and set sets. To achieve this and prevent data leakage, we use subject-wise K-fold cross-validation, guaranteeing that every subject in the VitalDB dataset serves as part of a test set exactly once across the folds. The reported performance is then the aggregate over all these test sets, providing a fair comparison. We train the model  for 50 epochs with a learning rate of 0.0003 and early stopping on the validation set. 
\paragraph{PPG Windkessel:} We train a density estimator to predict cardiovascular parameters on a dataset of in-silico PPGs (see \Cref{apd:exp_details}). As estimator we use the same NPE model architecture used for the other methods, and keep a $20\%$ validation split for model selection. Performance is evaluated on the VitalDB dataset. Compared to the other models, where longer training did not result in an improvement, for this baseline we observed improved performance when training for a larger number of epochs so we report the results for the model trained for 250 epochs with a learning rate of 0.0003, with early stopping on the validation set. 

\subsection{Metrics and Evaluation}

Following recent work \citep{Manduchi2024Leveraging} we focus our evaluation on assessing the ability of a model to predict temporal trends in cardiovascular parameters. To this end, we compare the predicted and ground truth values of a given cardiovascular biomarker for a given subject, over the time frame the subject is monitored. We compute the Sperman's correlation between the two time series. 
To mitigate short-term fluctuations and noise, both the predicted and ground truth time series are smoothed using an exponential moving average with a window size of 16 segments (i.e., 128 seconds), prior to computing correlation.

\section{Additional qualitative and quantitative results}

\subsection{Generative modeling for PPG-to-APW mapping}\label{app:gen_model}
Given the complexity of the cardiovascular system, the mapping between a PPG and the corresponding APW is complex and influenced by unobservable factors, introducing inherent ambiguities. Therefore, a single PPG segment may correspond to multiple physiologically plausible APWs. Consequently, training a deterministic model to map PPGs to APWs using a standard regression loss (e.g., Mean Squared Error) may be suboptimal, due to forcing a single averaged solution that fails to capture the variation in plausible generations. A generative model on the other hand may be a more sensible solution, with the possibility of capturing a distribution of plausible APWs through multiple generations. Therefore, in this work we employed a conditional VAE to model the conditional distribution of physiologically plausible APWs given an input PPG. To empirically validate this choice, in this section we compare the APW generations obtained with our approach, against the ones obtained with a neural network trained for direct APW reconstruction with an MSE loss.

We repeat the experiment described in \Cref{sec:experiments}, where produced APWs from VitalDB PPG inputs, are used for inference of cardiovascular parameters with our NPE model trained on in-silico APWs. Performance is assessed via the Spearman correlation  between the ground truth and predicted values for HR, SV, CO and SVR. The results demonstrate a significant performance advantage for our generative model, confirming that a deterministic mapping lacks the expressive power to capture the physiological variations inherent in the PPG-to-APW relationship, thereby empirically validating our modeling choice.

\begin{table*}[htbp] 
    \centering
    \setlength{\tabcolsep}{4pt}
    \begin{tabular}{lcccc}
        \toprule
        \textbf{Model} & \textbf{HR} & \textbf{SV} & \textbf{CO} & \textbf{SVR} \\
        \midrule
        Ours (1D-CNN)  & 0.94        & 0.31        & 0.40        & 0.24         \\
        Ours (ConditionalVAE)   &  0.96  & 0.49 & 0.53 & 0.29  \\
        \bottomrule
    \end{tabular}
        \caption{Comparison between modeling the PPG-to-APW generation via a conditional VAE model vs a determinisitc 1-D CNN trained via MSE loss. Physiological relevance of the generated signals is assessed by evaluating the accuracy of cardiovascular parameter inference on the VitalDB dataset, using an NPE estimator trained in-silico. Average Spearman correlation between ground truth and predicted values per-subject for different cardiovascular parameters is reported.}
    \label{tab:vitaldb_spearman_correlation}
\end{table*}

\subsection{Rejection of unusable PPG segments}
\label{app:rejecting-unusable-segments}
\begin{figure}[htpb]
    \centering
    \subfloat[High predicted uncertainty.\label{fig:unc:a}]{%
        \includegraphics[width=0.85\textwidth]{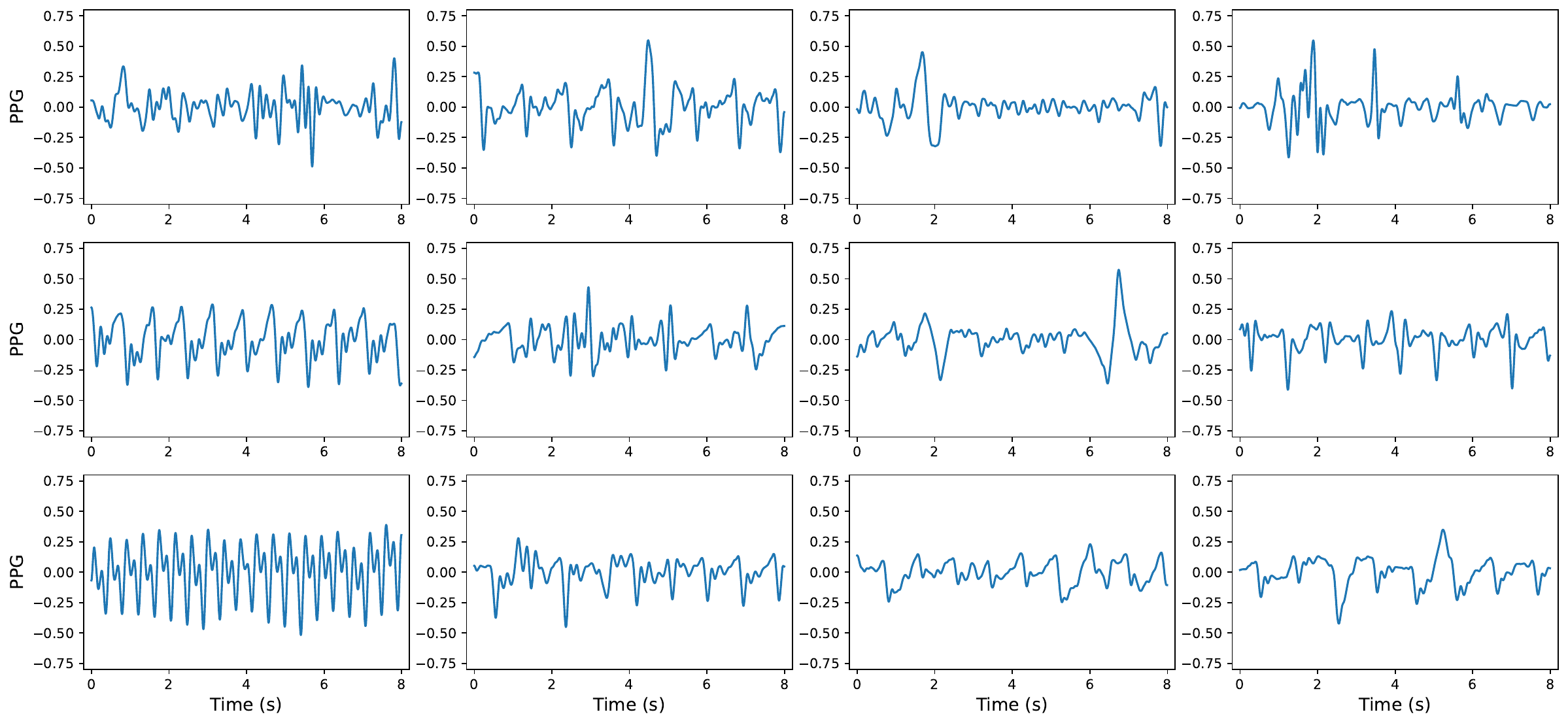}%
    } \\ 
    \subfloat[Low predicted uncertainty.\label{fig:unc:b}]{%
        \includegraphics[width=0.85\textwidth]{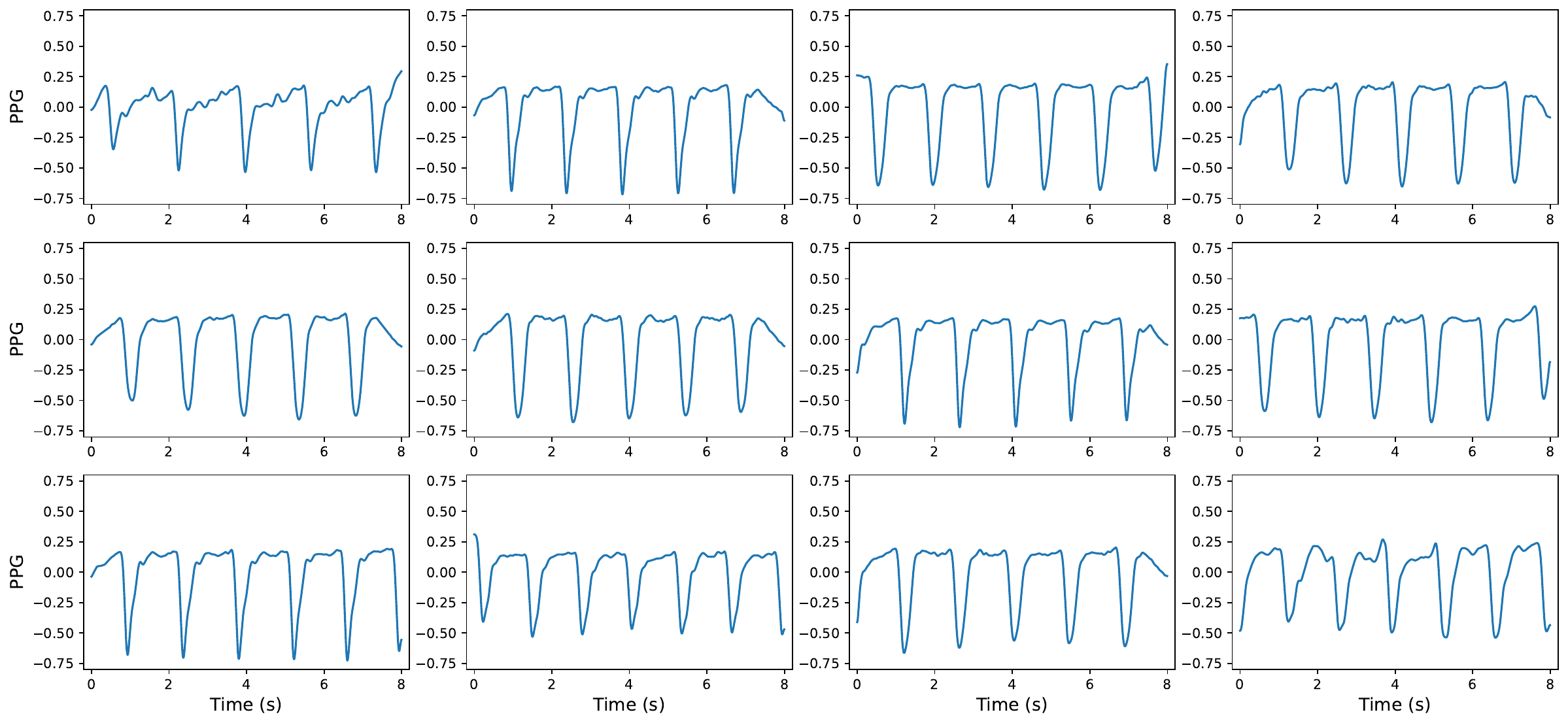}%
    }
    \caption{Rejecting unusable samples via predicted uncertainty for HR inference. (a) Twelve VitalDB PPG segments with the highest predicted uncertainty. (b) Twelve VitalDB PPG segments with the least predicted uncertainty. All PPG signals underwent bandpass filtering.}
    \label{fig:uncertainty-PPGs-combined}
\end{figure}

In this section we demonstrate the utility of uncertainty modeling in our cardiovascular parameter inference, through the rejection of unusable PPG measurements. In detail, we use the uncertainty estimator associated with the cardiovascular parameter prediction (see \Cref{sec:methodology}).  
In \Cref{fig:unc:a} we showcase segments with highest predicted heart rate (HR) uncertainty on the VitalDB dataset, demonstrating the ability to identify overly noisy input PPG samples for which inference is unreliable. \Cref{fig:unc:b} complements these results, showing that low predicted uncertainty corresponds to clean PPG segments.

\subsection{Absolute value prediction}
\Cref{tab:MAE-results} reports average per-subject Mean Absolute Error (MAE) for the compared models across the considered biomarkers. Our approach demonstrates superior performance to the PPG Windkessel model baseline, that also does not use labeled in-vivo data for training. However these results highlight that all approaches, including ours, currently exhibit significant limitations in absolute value prediction. This difficulty stems from the inherent dependence of complex hemodynamic parameters on unobservable, subject-specific physiological characteristics. Commercial devices often address this by integrating patient demographics \citep{hendy2016, manecke2005} or requiring an initial gold-standard assessment \citep{hendy2016}. In this work we focus on PPG-based cardiovascular assessment without resorting to in-vivo PPG signals with hemodynamic labels or patient-specific metadata for training or calibration. Consequently, within this specific setting, accurate absolute value prediction remains an elusive target. As discussed in \cref{sec:experiments}, personalization and calibration strategies represent promising directions to improve this performance aspect in future work.

\begin{table}[htbp]
    \centering
    \setlength{\tabcolsep}{4pt}
    \begin{tabular}{lcccc}
        \toprule
        \textbf{Model} & \textbf{HR} & \textbf{SV} & \textbf{CO} & \textbf{SVR} \\
        \midrule
        APW  & 0.962 (0.002)        & 0.653 (0.006)       & 0.725 (0.006)        & 0.341 (0.008)         \\
        PPG Windkessel  & 0.804 (0.040)        & 0.348 (0.035)    & 0.438 (0.027)        & 0.289 (0.020)         \\
        PPG Supervised  & 0.967 (0.001)        & 0.190 (0.010)        & 0.333 (0.020)        & 0.246 (0.006)         \\
        Ours   & 0.966 (0.005)  & 0.489 (0.003) & 0.531 (0.003)        & 0.288 (0.001)  \\
        \bottomrule
    \end{tabular}
        \caption{Per-subject Spearman correlation between ground truth and predicted values for four cardiovascular biomarkers, obtained with the approaches compared in \Cref{fig:cv-params-boxplots}. Results are averaged across three independent runs for each approach, and standard deviations are reported in parenthesis.
        }
    \label{tab:ext-num-results}
\end{table}

\vspace{10ex}
\begin{table}[htbp!]
    \centering
    \setlength{\tabcolsep}{4pt}
    \begin{tabular}{lcccc}
        \toprule
        \textbf{Model} & {\textbf{HR ($bpm$)}} & {\textbf{SV ($mL$)}} & {\textbf{SVR ($dyn \cdot s /{cm}^5$)}} & {\textbf{CO ($L/min$)}} \\
        \midrule
        APW            & 0.755               & 18.087              & 412.992              & 1.272               \\
        Ours           & 1.199               & 20.243              & 454.516              & 1.425               \\
        PPG Windkessel & 2.951               & 20.960              & 557.478              & 1.552               \\
        PPG Supervised & 0.725               & 18.789              & 328.499              & 1.329               \\
        \bottomrule
    \end{tabular}
        \caption{Average per-subject MAE results on the VitalDB dataset, for the models compared in \Cref{fig:cv-params-boxplots}. 
        }
    \label{tab:MAE-results}
\end{table}

\vspace{5ex}
\applefootnote{\textcolor{textgray}{\sffamily Apple and the Apple logo are trademarks of Apple Inc., registered in the U.S. and other countries and regions.}}

\end{document}